\RequirePackage{silence}
\documentclass[runningheads]{llncs}

\usepackage{xcolor}
\usepackage{graphicx}
\graphicspath{{/}{fig/}}
\usepackage{cite}
\usepackage{tikz}
\usepackage{array}
\usepackage{textcomp}
\usepackage{xcolor}
\usepackage{multirow}
\usepackage{booktabs}
\usepackage{float}
\usepackage{hyperref}

\usepackage{tikz}

\begin{document}
\title{An Overview of Federated Learning at the Edge and Distributed Ledger Technologies for Robotic and Autonomous Systems}
\author{
    Yu Xianjia \and
    Jorge Pe\~na Queralta \and
    Jukka Heikkonen \and \\
    Tomi Westerlund
}%
\authorrunning{Yu Xianjia et al.}
\titlerunning{Federated Learning in Robotic and Autonomous Systems}
\institute{
    %     School of Information Science and Technology \\
    %     Fudan University, China \\
    %     \email{\{qingqingli16, zhuo\}@fudan.edu.cn}\\
    % \and
        \href{https://tiers.utu.fi}{Turku Intelligent Embedded and Robotic Systems Lab, University of Turku, Turku, Finland} \\
        % University of Turku, Finland\\ 
        \email{\{xianjia.yu, jopequ, jukhei, tovewe\}@utu.fi} \\[+0.42em]
        \url{https://tiers.utu.fi}
    }
\maketitle
%
%
%
%%%%%%%%%%%%%%%%%%%%%%%%%%%%%%%%%%%%%%%%%%%%%%
%%                                          %%
%%                SECTIONS                  %%
%%                                          %%
%%%%%%%%%%%%%%%%%%%%%%%%%%%%%%%%%%%%%%%%%%%%%%
\begin{abstract}
% Autonomous systems are becoming progressively ubiquitous. Computation is shifting towards more distributed architectures at the edge, novel mobile connectivity solutions are enabling low-latency offloading and real-time collaboration and decentralized technologies are being increasingly adopted, with blockchain and distributed ledger technologies (DLTs) playing a key role. 
Autonomous systems are becoming inherently ubiquitous with the advancements of computing and communication solutions enabling low-latency offloading and real-time collaboration of distributed devices. Decentralized technologies with blockchain and distributed ledger technologies (DLTs) are playing a key role.
At the same time, advances in deep learning (DL) have significantly raised the degree of autonomy and level of intelligence of robotic and autonomous systems. While these technological revolutions were taking place, raising concerns in terms of data security and end-user privacy has become an inescapable research consideration. Federated learning (FL) is a promising solution to privacy-preserving DL at the edge, with an inherently distributed nature by learning on isolated data islands and communicating only model updates. However, FL by itself does not provide the levels of security and robustness required by today's standards in distributed autonomous systems. This survey covers applications of FL to autonomous robots, analyzes the role of DLT and FL for these systems, and introduces the key background concepts and considerations in current research.

%In the era of IoT, network-connected robots and autonomous devices are becoming progressively ubiquitous. Cloud and edge-based robots and autonomous systems have penetrated various aspects of our life including diverse private scenarios. Due to massive demands on intelligence and efficiency of these systems, there is a proliferation of robot learning methods such as deep learning, reinforcement learning, and imitation learning based on distributed data sets from multi-devices (isolated data islands). However, communication, security, and privacy issues are the major hurdles of the collaboration and learned knowledge sharing among those devices. Federated Learning (FL) as an effective and promising scheme to tackle data silos and data-sensitive issues has been proposed. Instead of sharing local data with the centralized cloud server for training a global model, FL aggregates the local models in the cloud server without the local data transmissions. This survey explores emerging opportunities of FL for robotic and autonomous systems. In this study, we contribute to concluding the recent development of FL at the edge and Federated Reinforcement Learning(FRL) in particular for their crucial roles in robotics and autonomous systems. Additionally, the recent applications of FL in Robotic and Autonomous Systems will be illustrated as well. 
% The major technologies including cloud and fog robotics, distributed learning, FL on edge devices, and evolution of FL that result in the integration of FL with robotic and autonomous systems will be introduced as well. 


\end{abstract}

\section{Introduction}
\label{sec:intro}

With a staggering increase in the number of connected devices being deployed worldwide within the Internet of Things (IoT), the amount of data that is generated and transmitted has grown at exponential rates. The inefficiency of processing all this data in a centralized manner at the cloud has brought forward new computing and networking paradigms in recent years~\cite{shi2016edge}. Computing at the edge, closed to where the data sources are, has evident benefits in terms of latency and bandwidth savings~\cite{queralta2019edge}. Another key advantage is the inherent benefits to data privacy, as raw data does not travel too far~\cite{metwaly2019edge}. At the same time, the data is being fed to increasingly complex artificial intelligence (AI) models, with deep learning (DL) in particular becoming pervasive across multiple fields and application domains. Recent years have also brought an increasing awareness to the risks and drawbacks of sharing personal data over the internet. The solution to distributed computing at the edge while preserving privacy of data and leveraging DL solutions is federated learning~\cite{yang2019federated}. Federated learning (FL) enables distributed training of complex models over isolated data islands from remote nodes (data sources). The local training results (updates to local models) are then aggregated, e.g. in a cloud server, and a global generalized model is shared back to the nodes. All this with zero raw data transmission~\cite{liu2020federated}.

From the perspective of robotic and autonomous systems, which are becoming increasingly ubiquitous%~\cite{ifr}
, cloud solutions have enabled higher degrees of intelligence by eliminating constraints of onboard computational and storage resources~\cite{kehoe2015survey}. Cloud robotics and AI robotics are now an essential part of state-of-the-art robotic systems. Furthermore, as mobile connectivity evolves, 5G and beyond networks are set to further bring the integration of AI, robotics and distributed networking solutions~\cite{bogue2017cloud}. Applications of AI in robotics include, e.g., the deployment of DL for natural language processing (NLP)~\cite{matuszek2018grounded}, computer vision~\cite{ruiz2018survey}, or in navigation and mapping~\cite{li2019deep}. In control, Reinforcement learning (RL) has been successfully applied in complex games~\cite{shao2019survey} and its relevance for dexterous manipulation extensively demonstrated~\cite{rajeswaran2017learning}. Deep reinforcement learning (DRL) is particularly relevant to autonomous robots~\cite{henderson2018deep}.

Federated learning provides a framework for more efficient learning in distributed autonomous systems and multi-robot systems, enabling collaborative learning across heterogeneous cloud and edge nodes and a wide range of robotic and autonomous systems. Even though cloud robotics can promote scalability, collaborative knowledge sharing, and development operation of robotics and autonomous systems, challenges remain in, e.g., security or reliable connectivity. %% FOR ARXIV AFTER MODIFICATION

\begin{figure}
    \centering
    \includegraphics[width=\textwidth]{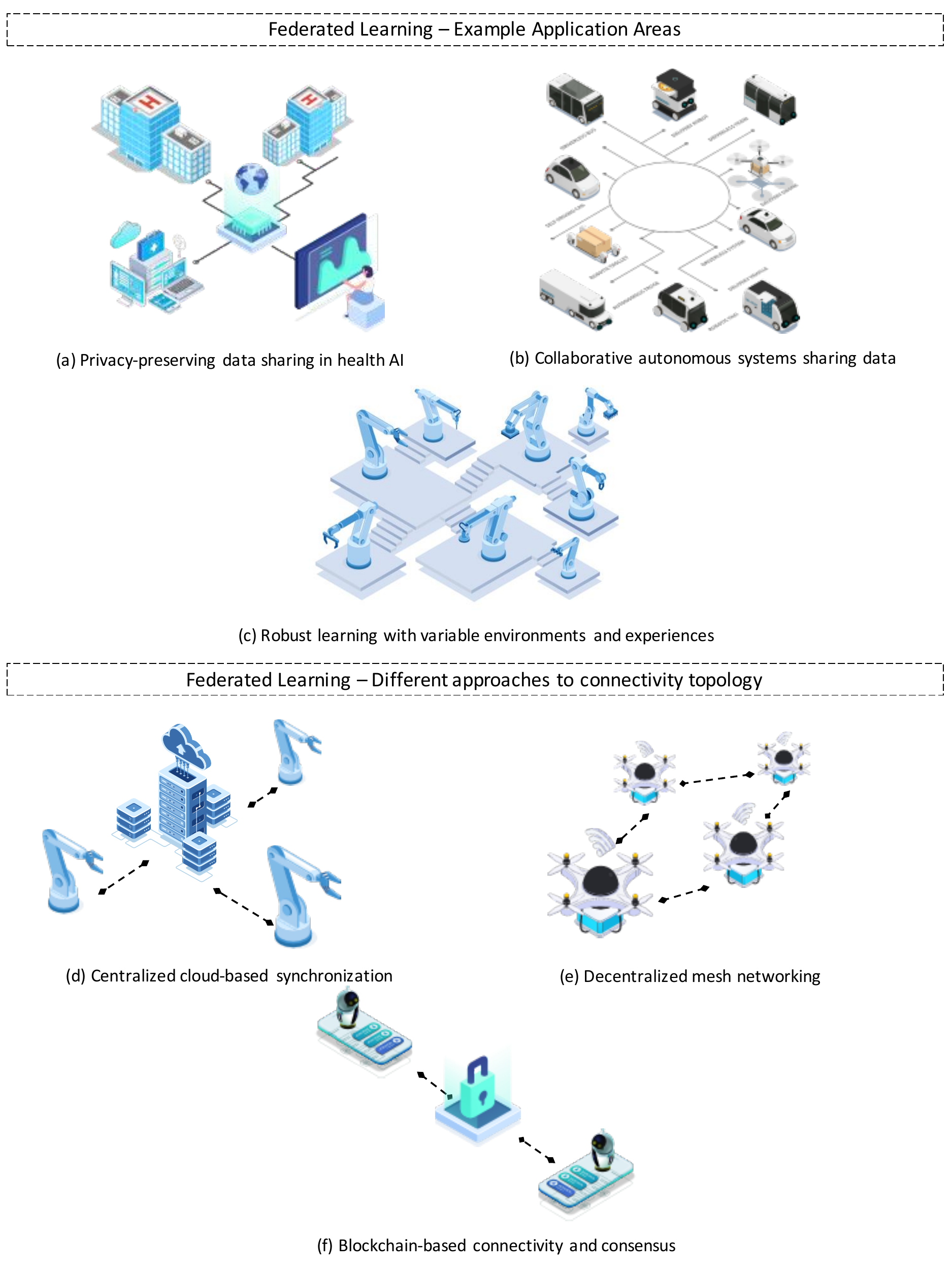}
    \caption{Conceptual illustration showing potential application areas and connectivity topologies in federated learning systems.}
    \vspace{-1.42em}
    \label{fig:concept}
\end{figure}

Multiple reviews and survey papers in the literature have been devoted to studying design approaches, implementation details and application possibilities of FL. Compared to current works focused on security and privacy~\cite{mothukuri2021survey}, personalized FL~\cite{kulkarni2020survey}, or communication at the edge~\cite{lim2020federated}, the present work aims to provide a comprehensive view of how FL can be leveraged to raise the level of autonomy and degree of intelligence of robotic systems. We look at different application opportunities at the edge and within autonomous mobile robots. We provide an overview of the most important concepts, and pay particular attention to synergies between FL and distributed ledger technologies (DLTs), among which blockchain technology has gained significant attention. A conceptual illustration of FL applications and approaches to connectivity is shown in Figure~\ref{fig:concept}.

% Owing to the flourish of the Internet of things (IoT) in recent years, voluminous amounts of data have been generated including sensitive and private data and a large number of edge devices have been connected by networks. As an important part of IoT, robotic and autonomous systems have been deployed in a largely growing number~\cite{ifr}. Cloud robotic and autonomous systems have made the limitation of onboard resources such as computation and storage no longer exist via data sharing and multi-robot cooperation through networks~\cite{kehoe2015survey}. Thanks to the flexible provisioning supported by 5G~\cite{bogue2017cloud}, cloud infrastructure as a service provides great convenience for design small and decoupled components such as multi robots using off-the-shelf tooling.

The rest of this survey is organized as follows. In Section II, we introduce key concepts and background to FL frameworks and related research areas. Section III explores the synergies with blockchain and other DLTs. Then, Section IV shifts the focus towards FL at the edge, and its significance and relevance to robotic and autonomous systems. Section V overviews application domains, and the work is concluded in Section VI. % FOR ARXIV

% novel techniques and recent research trends about robotic and autonomous systems. In section III, Federated reinforcement learning (FRL) is further studied in particular given that reinforcement learning is significantly relevant to robotic and autonomous systems. Then, Section IV presents the work about FL in edge devices which are extensive applied in robotic and autonomous systems. We discuss the main trends and open problems in Section V, with Section VI concluding the work.

\section{Background}
\label{sec:bg}

The adoption and development of FL frameworks have been directly or indirectly influenced by other technological and paradigm trends in robotics and autonomous systems. Since the invention of FL, there are a number of works on optimization of FL itself. Different research directions include increasing the adaptiveness, enhancing the privacy-preserving properties, or building towards more efficient collaboration for distributed robot learning, among others. In this section, we briefly introduce the different identifiable research directions from the literature, and the concepts that underpin the popularity of FL in robotics and autonomous systems.

\subsection{Cloud Robotics and Automation}

Cloud robotics is a field of robotics that capitalizes on cloud technologies.
The cloud infrastructure can provide robots and autonomous systems with extensive resources and potential benefits including big data, cloud computing, collective robot learning, and human learning~\cite{kehoe2015survey}. Under cloud infrastructures, robotic systems have access to more collaborative approaches to autonomy, faster processing of deep learning models, and more powerful computational capabilities in general. A collection of robots in different areas or states can cooperate in a variety of tasks such as disaster management identifying several critical challenges~\cite{chen2018study} and manufacturing environment~\cite{yan2017cloud}. There are a number of examples and implementations of cloud robotics platforms such as AWS RobotMaker~\cite{robo_maker} from Amazon, Dex-Net~\cite{dex-net} from UC Berkeley Automation Lab,  and Google cloud robotics~\cite{goolge-cloud-rob}. % FOR ARXIV

\subsection{Distributed Deep Learning}

With the increasing amount of data and complexity of DL models, the process of training models becomes inherently costly, computation-intensive, and time-consuming. Distributed DL was proposed to utilize the  multiple processors to accelerate DL training process by parallel the computation and the data~\cite{xing2015petuum, tang2020communication, assran2019stochastic}. There is a significant amount of work in the literature dedicated to distributed DL in the pursuit of closer collaboration between cloud and edge computing~\cite{wu2020collaborate, jiang2019distributed, chen2019exploring}. This balance between the two paradigms is set to become increasingly pervasive with a well-established IoT era. Immediate concerns that raise with the deployment of distributed DL across cloud and edge is the security of data and privacy of users. In consequence, multiple research directions have emerged to  make distributed learning processes more scalable, secure and privacy-preserving through~\cite{shi2021towards, li2020toward, buniatyan2019hyper, duan2020privacy, zhang2019privacy}. Additionally, other research efforts are directed towards utilizing distributed DL for processing and learning from sensitive data such as health data~\cite{vepakomma2019reducing}, video surveillance data ~\cite{chen2019distributed} and medical data from multiple private or public institutions~\cite{balachandar2020accounting}.

\subsection{Privacy and Security in Deep Learning and Federated Learning}

With the wider adoption of DL over the past decade, issues regarding data security and privacy of data sources became increasingly studied. Some of the main types of security-related issues in DL appear with evasion attacks during model inference and poisoning attack during model training~\cite{bae2018security}. Adversarial attacks to the algorithms, and model reconstruction attacks are other examples. Multiple solutions have been proposed to deal with these and other attack vectors, including differential privacy, homomorphic encryption, data anonymization, pseudonymization, algorithm encryption, or hardware security implementations, among others~\cite{kaissis2020secure}. Despite the efforts, new attack vectors have appeared such as re-identification attacks (identification of individual data sources despite data anonymization techniques based on other information in the datasets), dataset reconstruction attacks, or tracing attacks (also referred to as membership inference, though which the inclusion of a specific individual in a dataset is inferred). While FL itself offers privacy-preserving attributes, the security robustness depends largely on the implementation and deployment methodologies. A recent survey on the topic~\cite{mothukuri2021survey} presents a comprehensive study on the current security and privacy concerning aspects with the conclusion that fewer privacy-specific threats than security-specific ones exist. Among these are, e.g., communication bottlenecks, poisoning, and backdoor attacks, especially inference-based ones.

\subsection{Federated and Distributed Reinforcement Learning}

Multi-agent RL is regarded as essential to realize general intelligence and cooperative environment learning. %~\cite{lanctot2017unified, da2019survey}. 
The main objective of a multi-agent RL is to obtain the localized policies and maximize the global reward for knowledge sharing on the premise of increased system complexity and computation~\cite{zhao2020sim}. In multi-robot systems, distributed RL can be leveraged to expose different robots to different environments, or to learn more robust policies in the presence of disturbances~\cite{zhao2020towards, zhao2020ubiquitous}.

While the literature in distributed RL is extensive, most works rely on sharing raw experiences or training in a centralized manner. Federated RL (FRL)~\cite{zhuo2019federated} has been proposed as an efficient solution for achieving high-quality policy transfer with the protection of both data and model privacy. FRL can be applied, e.g., to understand user behavior and adapt to it~\cite{nadiger2019federated}. In~\cite{lim2020federated}, FRL was proposed to allow multiple RL agents to learn optimal control policies for a series of IoT devices with slightly different dynamics. In another direction, FRL is regarded as an efficient method for resource allocation among networked devices~\cite{ruan2020low, nguyen2020resource}.

\subsection{Recent Trends in Federated Learning}

Federated learning has arguably raised the possibilities for collaborative learning across multiple independent agents. In this section, we give an overview of works that have focused towards improving specific aspects of FL.

With a focus on scalability, a high-level designed FL system based on TensorFlow has been developed that draws significant conclusions on existing challenges and future research directions~\cite{bonawitz2019towards}. From the perspective of system security, a systematic study of Byzantine-robust federated learning in~\cite{fang2020local} shows different approaches to  secure FL systems and make them more robust against local model poisoning attacks. A similar approach in~\cite{li2020learning}, instead considers a solution to detect the malicious model updates in every round of training process before aggregating the locals models in the centralized cloud server. Owing to a wide range of approaches relying on a the centralized cloud server for aggregation of local model updates, FL frameworks may fail if a malicious aggregation server takes over the central FL node. To cope with this problem, dispersed FL~\cite{khan2020dispersed} has been proposed, where a global model is yielded in either a centralized or distributed manner through the aggregation of sub-global models, which are iteratively computed based on different groups similar to traditional FL approaches.

Machine learning itself can also play a role in improving the performance of FL systems. In~\cite{lu2020blockchain}, deep reinforcement learning is used to select the optimized edge nodes and the learned model parameters are integrated into a blockchain-based FL scheme for enhanced security and reliability. Furthermore, combining with other privacy-preserving machine learning methods such as differential privacy~\cite{wei2020federated} and modern cryptography techniques such as homomorphic encryption~\cite{zhang2020batchcrypt}, FL can achieve high level privacy-preserving and secure capabilities.

It is also worth meaning at this point that FL solutions are specialized in aggregating local models to a global model for knowledge sharing. Nonetheless, in terms of the characterization of heterogeneous data collected across large-scale deployments of edge devices, it is often essential to the application to make the models discriminative in each device. In this direction, personalized FL was proposed to tackle the aforementioned problem by further performing a series of learning steps locally after receiving the global model from the cloud server, based mostly on locally available data for which the model needs to be tailored~\cite{deng2020adaptive, fallah2020personalized}.
\section{Federated Learning at the Edge}

Federated learning has emerged within the wider edge computing paradigm. Deploying FL at the edge has gained significant attention from the research community owing to the availability of rapidly increasing amounts of data and computational resources at the edge.  %After 2018, the studies became progressively extensive. The research mainly focused on task allocation, 
Research directions include the deployment FL in resource constrained embedded systems, communication-efficient FL, energy-efficient FL and privacy-preserving federated edge learning with the aim to improve the learning performance in networks where the general assumption is that resources are inherently at the edge~\cite{wang2019adaptive}. For instance, an early work explored how to capitalize on FL to optimize the caching scheme in edge computing process~\cite{yu2018federated}.

\subsection{Task Allocation}

A general problem in distributed systems is task allocation. Learning more efficient task allocation at the edge can produce more effective strategies for worker selection and load distribution. Doing so through a distributed FL framework is a natural fit to such systems. In~\cite{kang2020training}, a matching-theoretic approach was proposed for task assignments schemes in federated edge learning framework to solve the task assignment problem between the workers and multiple task publishers with efficient performance. In another work, an asynchronous task allocation method was introduced to realize equal task allocation within the FL system itself, i.e., minimizing the maximum difference between the number of model updates done by every worker in a FL edge network~\cite{mohammad2019adaptive}.

% @inproceedings{dong2020offloading,
%   title={Offloading Federated Learning Task to Edge Computing with Trust Execution Environment},
%   author={Dong, Shifu and Zeng, Deze and Gu, Lin and Guo, Song},
%   booktitle={2020 IEEE 17th International Conference on Mobile Ad Hoc and Sensor Systems (MASS)},
%   pages={491--496},
%   year={2020},
%   organization={IEEE}
% }

\subsection{Communication and Energy Efficiency}

Multiple studies in the literature focus on mitigating the bottleneck that communication latency can become in FL systems. Some of the proposed solutions involve the aggregation over the air of multiple updates  from an analog perspective, rather than relying on conventional orthogonal network access~\cite{zhu2018broadband}. In a similar direction, and to mitigate the communication overhead, authors in~\cite{lu2020communication} introduced an asynchronous communication model for digital twin edge networks. In their work, FL is formulated as an optimization problem that aims at reducing the communication cost by decomposing it and using DNN for communication resource allocation. In~\cite{mills2019communication}, compression techniques were utilized to realize a more communication-efficient FL solution.

Regarding to energy efficiency of FL, authors in~\cite{li2020talk} tackled the problem of improving the energy efficiency of FL by developing a convergence-guaranteed algorithm with flexible communication compression. In~\cite{mo2020energy}, two transmission protocols based on the non orthogonal multiple access and time division multiple access were considered to jointly optimizing the transmission power and rate at edge devices in a federated edge learning system. Other authors showed that learning an optimal resource-management policy substantial energy can be reduced in a FL system~\cite{zeng2020energy}.

\subsection{Client Selection}

Client selection is a process to choose model updates from certain clients to be aggregated, especially when computational resources are constrained and complex aggregation processes are not possible. In~\cite{nishio2019client}, a framework named FedCS (Federated Client Selection) was introduced to dynamically select and maximize the number of clients (training agents) in heterogeneous edge networks. The dynamic approach was based on an online estimation of actively available resources. The results show that such approach can provide significant better training performance with heterogeneity of resources across clients, with overall significantly shorter training times than traditional FL methods. In another work, an optimization algorithm is designed to jointly optimize the data sampling and user selection strategies, which is shown to approach to the stationary optimal solution efficiently~\cite{feng2020joint}.

\subsection{Privacy-Preserving and Secure Mechanisms}

While FL is flexible in nature and inherently deals with issues related to data ownership and governance, it does guarantee privacy and security by itself. Integration of other techniques and approaches to data security and user privacy needs to be considered to achieve a robust FL framework. For instance, an asynchronous FL system~\cite{lu2020privacy} with the incorporation of local differential privacy for enhanced privacy of local modes updates has been proposed in the literature~\cite{lu2019differentially}. 
To tackle the problem of active poisoning attacks, which FL is vulnerable to, authors in~\cite{zhang2020poisongan} generated a model for different poisoning attacks based on generative adversarial networks (GANs). Utilizing GANs, which is a well established approach in DL research, opens the door to more robust FL systems.

\section{Synergies between Federated Learning and Distributed Ledger Technologies}

Distributed ledger technologies have multiple applications in multi-robot systems and distributed autonomous systems. Blockchain technology, in particular, has been applied to robot swarms able to deal with byzantine agents~\cite{ferrer2018blockchain}, for sharing computational and communication resources~\cite{queralta2019blockchain}, but also for privacy-critical applications~\cite{nawaz2019edge, nawaz2020edge}. The distributed consensus algorithms in DLTs, the auditability of operations and the built-in encryption, among others, aid in designing more secure and privacy-preserving systems at the edge~\cite{queralta2020blockchain}. Blockchain technology and subsequent DLT solutions can be thus leveraged as the basis for trust and credibility in a distributed system.

Traditional FL approaches rely on a centralized cloud server for model aggregation, therefore assuming such central node has full trust from the rest of the system. In practice, the reliance on the cloud server and the transmission to the local clients can be threatened by various types malicious attacks. Additionally, the scalability of the system is inherently limited by the existence of the single processing node. Even if it is replicated in the cloud, there is still a strong reliance on trusted cloud servers. Therefore, being able to deploy trustable FL frameworks in a distributed and decentralized manner can take FL to new application domains~\cite{nguyen2021federated, bao2019flchain, majeed2019flchain}.

The literature on applications of DLTs and FL for robotic systems is sparse. At present, studies on applications of blockchain-enhanced FL mainly focus on autonomous vehicles and the Internet of Vehicles (IoVs). The core objective of these is studies is to build a trustworthy vehicular network without any centralized training process or trusted third party. In this direction, 
blockchain-supported FL has been proposed to build a trustworthy vehicular network. with performance metrics including accuracy, energy consumption, and lifetime rate, along with throughput and latency evaluated by simulation~\cite{otoum2020blockchain}. It is worth mentioning as well that a hierarchical blockchain based FL has also proved to be efficient in building towards large-scale vehicular networks and shown potential resilience against certain malicious attacks~\cite{chai2020hierarchical}. In another work, an autonomous blockchain enabled FL has been proposed to add further privacy-preserving properties and efficient local on-vehicle machine learning model aggregation in a decentralized manner~\cite{pokhrelfederated}. The authors indicate some key challenges of the proposed framework in the autonomous vehicles field including sophisticated mobility models, mobility-aware and efficient verification, or privacy leakage risk analysis.

Other examples of blockchain-enhanced FL include drones in 6G networks and control in railway systems. In~\cite{gaofeng2020blockchain}, with the objective of replacing the manual fraction and braking operations with automatic operations in a heavy haul railway system,  blockchain-based  FL was utilized to obtain a novel ML model for intelligent control under the circumstance of the imbalanced fraction and braking data. % In buildings towards new systems
One approach to build the foundations for the upcoming 6G era, a blockchain-based empowered FL with the applications of mobile miners at drones has been proposed for a disaster response system~\cite{pokhrel2020federated}. In this work, the authors mainly focused on the definition of frameworks and analysis of blockchain latency and energy consumption.
\section{Applications of FL In Robotic and Autonomous System}

Networked robotic and autonomous systems are becoming ubiquitous. These agents are in turn increasingly heterogeneous in terms of their computational capabilities but also the type of data they produce. With the wider availability of unprecedented amounts of data, deep learning has been broadly employed across autonomous robots of all types. Cloud robotics unlocks for robotic and autonomous systems access to potentially unlimited computational, memory or storage resources, partially avoiding the limitations of onboard resources. Cloud robotics also offers the use of the internet for massive parallel computing and resource sharing~\cite{goldberg2013cloud}. At the same time, autonomous robotic solutions have been adopted across a growing number of industries and application domains. These include data-sensitive scenarios such as hospitals, military bases or hotels. On account of features ranging from privacy preservation, decentralized reliability, minimal communication and focus on on-board computation, it is arguable that federated learning has potential to be a secure and efficient robot learning framework in and it will be further adopted across different types of autonomous systems~\cite{savazzi2021opportunities}.

% \begin{figure}[H]
%     \centering
%     \includegraphics[width=0.90\textwidth]{fig/A_FL_new.png}
%     % \includegraphics[width=0.8\textwidth]{figs/Application_FL.pdf}
%     \caption{Applications of Federated Learning in Robotic and Autonomous Systems}
%     \label{fig:robots}
% \end{figure}

% \subsection{Cloud and Fog Robotics}
% FL has its potential as a secure and privacy-preserving framework for deep robot learning such as imitation learning, life-long reinforcement learning in cloud  robotics with heterogeneous sensor data inputs~\cite{liu2019federated, liu2019lifelong}.

From a system-level perspective, different nomenclatures are used to define the paradigm of shared computation across and between cloud and edge. In this area, fog robotics has been introduced as the paradigm of deploying robot deep learning across shared computational, storage, and networking resources between cloud and edge in a federated way. In~\cite{tanwani2019fog}, the authors evaluated the performance of the designed fog robotics system through a surface decluttering application with object recognition approaches. They trained the deep models in the cloud server based on the non-private images, adapted and deployed the model based on the real world images on the edge side to reduce the round-trip communication cost. In another application of fog robotics, blockchain-based FL has been proposed for autonomous vehicles which enables a communication network where on-vehicle machine learning model are verified and exchanged in a distributed and privacy-aware fashion~\cite{pokhrel2020federated}. The authors evaluate the performance of generation delay, block propagation, and upload-download delay, showing promising applications of such frameworks.

% In a multi-robot system, differential federated learning is a solution for real-time data processing since unprecedented amount of data has been generated because of the ubiquitous robots and IoT services~\cite{zhou2018real}. In distributed multiple mobile robots integrated with FL, by monitoring client activities and leveraging available local computing resources, the learning process among the resource-constrained robots can be accelerated~\cite{imteaj2020fedar}.

% For developing specific robots, FL has been taken into account at the beginning of the system designing. In the process of designing a depression treatment robot, FL has been integrated in the system framework incorporating deep learning methods for robot psychotherapy to make the patient treatment more intelligent, efficient, and privacy-preserving~\cite{liu2021federated}. In other scenarios using cellular-connected UAVs to detect the outage in the environment, FL is applied to build a global model of the outage probability and optimize the UAVs' paths since each UAV covers limited region information~\cite{khamidehi2020federated}.

% Regarding specific field of robotic system, FL has proved to be effective in promoting their efficiency and  securing the data safety. 
Federated learning has potential within multiple specific autonomy problems and robotic subsystems. In~\cite{li2019fc}, cooperative SLAM based on visual-Lidar have been proposed by deploying a federated deep learning algorithm for feature extraction and dynamic map fusion without transferring original images among the robots. In the area of dynamic map fusion, authors in~\cite{zhang2021distributed} developed a novel fusion scheme among the networked vehicles supported by FL. Superior performance and robustness were then demonstrated in the Car Learning to Act (CARLA) simulation platform. In~\cite{majcherczyk2020flow}, trajectories forecasting (spatio-temporal predictions) has been performed in a multi-robot system through different FL variants: traditional FL approach where a cloud server aggregates the local models and serverless version. In the paper, the authors found that in a trajectories forecasting task, the results of the above methods are not notably different and they provided the first federated learning dataset obtained from multi-robot behaviors. FL has also proven to be an efficient and novel framework in heterogeneous sensor data fusion for imitation learning~\cite{liu2020federated02}. In terms of situational awareness, continuous learning has been demonstrated to be feasible through FL as a framework across computationally limited edge devices while enabling the post-deployment of learned models in inference-only mode~\cite{busart2020federated}. 
In~\cite{lim2020federated}, FRL was applied to learn an optimal control policy among multiple IoT autonomous devices of the same type.
In~\cite{liang2019federated}, the authors introduced a FL-based online reinforcement transfer learning process for real-time perception, with a demonstration through a collision avoidance system simulated in Airsim. From a more general autonomous navigation perspective, planning modules in cloud robotic systems can utilize federated reinforcement learning as a learning architecture for fusing prior knowledge and quickly adapting to new environments~\cite{liu2019lifelong}.

In the area of human-robot collaborative learning, a novel cognitive architecture based on FL was introduced for multi-agent learning from demonstration (LfD) with multiple humans incorporated in the self robot learning loop~\cite{papadopoulos2020towards}. In a subsequent study, the authors integrated the short- and long-term analysis of human behavior within their cognitive robot learning architecture to show that it can adaptively enhance large-scale multi-agent LfD~\cite{papadopoulos2021user}.
\section{Conclusion}
FL offers advantageous solutions to collaborative learning in decentralized multi-robot systems and distributed autonomous systems. FL will play a key role in networked ubiquitous robots and autonomous intelligent systems at the edge.  The vast and rapidly growing amount of research in the area is revealing the efficiency and applicability of FL in various solutions.The key advantages of FL solutions include the optimization of networking resources, resilience through decentralization, and inherent privacy-preserving properties by processing data directly at the edge.
We have also reviewed DLT-empowered FL, with DLTs in general having drawn significant attention in the robotics domain in recent years. DLT solutions, and blockchain technology in particular, can be the backbone of decentralized local model aggregation in a more privacy-preserving, secure, and distributed manner. Some of the most prominent results are being shown in the era of the internet of vehicles, set to become increasingly important with the wider adoption of 5G and beyond mobile connectivity solutions.
In summary, FL has multiple application possibilities in autonomous systems, either from a system-level perspective or within specific subsystems in, e.g., autonomous robots. Key research directions that need further exploration include optimization of communication, energy-efficiency at the edge, personalized FL, and further privacy and security enhancements. Research efforts are currently capitalizing on multidisciplinary approaches including modern encryption, novel connectivity topologies or new learning paradigms.

\bibliographystyle{unsrt}
\bibliography{bibliography}

\begin{thebibliography}{10}

\bibitem{shi2016edge}
Weisong Shi, Jie Cao, Quan Zhang, Youhuizi Li, and Lanyu Xu.
\newblock Edge computing: Vision and challenges.
\newblock {\em IEEE internet of things journal}, 3(5):637--646, 2016.

\bibitem{queralta2019edge}
Jorge {Pe\~{n}a Queralta}, Tuan~Nguyen Gia, Hannu Tenhunen, and Tomi
  Westerlund.
\newblock Edge-ai in lora-based health monitoring: Fall detection system with
  fog computing and lstm recurrent neural networks.
\newblock In {\em 2019 42nd international conference on telecommunications and
  signal processing (TSP)}, pages 601--604. IEEE, 2019.

\bibitem{metwaly2019edge}
Aly Metwaly, Jorge {Pe{\~n}a Queralta}, Victor~Kathan Sarker, Tuan~Nguyen Gia,
  Omar Nasir, and Tomi Westerlund.
\newblock Edge computing with embedded ai: Thermal image analysis for occupancy
  estimation in intelligent buildings.
\newblock In {\em Proceedings of the INTelligent Embedded Systems Architectures
  and Applications Workshop 2019}, pages 1--6, 2019.

\bibitem{yang2019federated}
Qiang Yang, Yang Liu, Tianjian Chen, and Yongxin Tong.
\newblock Federated machine learning: Concept and applications.
\newblock {\em ACM Transactions on Intelligent Systems and Technology (TIST)},
  10(2):1--19, 2019.

\bibitem{liu2020federated}
Yi~Liu, Xingliang Yuan, Zehui Xiong, Jiawen Kang, Xiaofei Wang, and Dusit
  Niyato.
\newblock Federated learning for 6g communications: Challenges, methods, and
  future directions.
\newblock {\em China Communications}, 17(9):105--118, 2020.

\bibitem{kehoe2015survey}
Ben Kehoe, Sachin Patil, Pieter Abbeel, and Ken Goldberg.
\newblock A survey of research on cloud robotics and automation.
\newblock {\em IEEE Transactions on automation science and engineering},
  12(2):398--409, 2015.

\bibitem{bogue2017cloud}
Robert Bogue.
\newblock Cloud robotics: a review of technologies, developments and
  applications.
\newblock {\em Industrial Robot: An International Journal}, 2017.

\bibitem{matuszek2018grounded}
Cynthia Matuszek.
\newblock Grounded language learning: Where robotics and nlp meet (invited
  talk).
\newblock In {\em Proceedings of the International Joint Conference on
  Artificial Intelligence}, 2018.

\bibitem{ruiz2018survey}
Javier Ruiz-del Solar, Patricio Loncomilla, and Naiomi Soto.
\newblock A survey on deep learning methods for robot vision.
\newblock {\em arXiv preprint arXiv:1803.10862}, 2018.

\bibitem{li2019deep}
Haoran Li, Qichao Zhang, and Dongbin Zhao.
\newblock Deep reinforcement learning-based automatic exploration for
  navigation in unknown environment.
\newblock {\em IEEE transactions on neural networks and learning systems},
  31(6):2064--2076, 2019.

\bibitem{shao2019survey}
Kun Shao, Zhentao Tang, Yuanheng Zhu, Nannan Li, and Dongbin Zhao.
\newblock A survey of deep reinforcement learning in video games.
\newblock {\em arXiv preprint arXiv:1912.10944}, 2019.

\bibitem{rajeswaran2017learning}
Aravind Rajeswaran, Vikash Kumar, Abhishek Gupta, Giulia Vezzani, John
  Schulman, Emanuel Todorov, and Sergey Levine.
\newblock Learning complex dexterous manipulation with deep reinforcement
  learning and demonstrations.
\newblock {\em arXiv preprint arXiv:1709.10087}, 2017.

\bibitem{henderson2018deep}
Peter Henderson, Riashat Islam, Philip Bachman, Joelle Pineau, Doina Precup,
  and David Meger.
\newblock Deep reinforcement learning that matters.
\newblock In {\em Proceedings of the AAAI Conference on Artificial
  Intelligence}, volume~32, 2018.

\bibitem{mothukuri2021survey}
Viraaji Mothukuri, Reza~M Parizi, Seyedamin Pouriyeh, Yan Huang, Ali
  Dehghantanha, and Gautam Srivastava.
\newblock A survey on security and privacy of federated learning.
\newblock {\em Future Generation Computer Systems}, 115:619--640, 2021.

\bibitem{kulkarni2020survey}
Viraj Kulkarni, Milind Kulkarni, and Aniruddha Pant.
\newblock Survey of personalization techniques for federated learning.
\newblock In {\em 2020 Fourth World Conference on Smart Trends in Systems,
  Security and Sustainability (WorldS4)}, pages 794--797. IEEE, 2020.

\bibitem{lim2020federated}
Wei Yang~Bryan Lim, Nguyen~Cong Luong, Dinh~Thai Hoang, Yutao Jiao, Ying-Chang
  Liang, Qiang Yang, Dusit Niyato, and Chunyan Miao.
\newblock Federated learning in mobile edge networks: A comprehensive survey.
\newblock {\em IEEE Communications Surveys \& Tutorials}, 22(3):2031--2063,
  2020.

\bibitem{chen2018study}
Wuhui Chen, Yuichi Yaguchi, Keitaro Naruse, Yutaka Watanobe, Keita Nakamura,
  and Jun Ogawa.
\newblock A study of robotic cooperation in cloud robotics: Architecture and
  challenges.
\newblock {\em IEEE Access}, 6:36662--36682, 2018.

\bibitem{yan2017cloud}
Hehua Yan, Qingsong Hua, Yingying Wang, Wenguo Wei, and Muhammad Imran.
\newblock Cloud robotics in smart manufacturing environments: Challenges and
  countermeasures.
\newblock {\em Computers \&amp; Electrical Engineering}, 63:56--65, 2017.

\bibitem{robo_maker}
AWS.
\newblock Aws robomaker.
\newblock \url{https://aws.amazon.com/robomaker/}.
\newblock [Online] - Last access: 2021-04-09.

\bibitem{dex-net}
AUTOLAB: UC Berkeley~Automation Lab.
\newblock Dex-net.
\newblock \url{https://berkeleyautomation.github.io/dex-net/}.
\newblock [Online] - Last access: 2021-04-09.

\bibitem{goolge-cloud-rob}
Google.
\newblock Google cloud robotics.
\newblock \url{https://googlecloudrobotics.github.io/core/}.
\newblock [Online] - Last access: 2021-04-09.

\bibitem{xing2015petuum}
Eric~P Xing, Qirong Ho, Wei Dai, Jin~Kyu Kim, Jinliang Wei, Seunghak Lee, Xun
  Zheng, Pengtao Xie, Abhimanu Kumar, and Yaoliang Yu.
\newblock Petuum: A new platform for distributed machine learning on big data.
\newblock {\em IEEE Transactions on Big Data}, 1(2):49--67, 2015.

\bibitem{tang2020communication}
Zhenheng Tang, Shaohuai Shi, Xiaowen Chu, Wei Wang, and Bo~Li.
\newblock Communication-efficient distributed deep learning: A comprehensive
  survey.
\newblock {\em arXiv preprint arXiv:2003.06307}, 2020.

\bibitem{assran2019stochastic}
Mahmoud Assran, Nicolas Loizou, Nicolas Ballas, and Mike Rabbat.
\newblock Stochastic gradient push for distributed deep learning.
\newblock In {\em International Conference on Machine Learning}, pages
  344--353. PMLR, 2019.

\bibitem{wu2020collaborate}
Huaming Wu, Ziru Zhang, Chang Guan, Katinka Wolter, and Minxian Xu.
\newblock Collaborate edge and cloud computing with distributed deep learning
  for smart city internet of things.
\newblock {\em IEEE Internet of Things Journal}, 7(9):8099--8110, 2020.

\bibitem{jiang2019distributed}
Haotian Jiang, James Starkman, Yu-Ju Lee, Huan Chen, Xiaoye Qian, and Ming-Chun
  Huang.
\newblock Distributed deep learning optimized system over the cloud and smart
  phone devices.
\newblock {\em IEEE Transactions on Mobile Computing}, 20(1):147--161, 2019.

\bibitem{chen2019exploring}
Yitao Chen, Kaiqi Zhao, Baoxin Li, and Ming Zhao.
\newblock Exploring the use of synthetic gradients for distributed deep
  learning across cloud and edge resources.
\newblock In {\em 2nd $\{$USENIX$\}$ Workshop on Hot Topics in Edge Computing
  (HotEdge 19)}, 2019.

\bibitem{shi2021towards}
Shaohuai Shi, Xianhao Zhou, Shutao Song, Xingyao Wang, Zilin Zhu, Xue Huang,
  Xinan Jiang, Feihu Zhou, Zhenyu Guo, Liqiang Xie, et~al.
\newblock Towards scalable distributed training of deep learning on public
  cloud clusters.
\newblock {\em Proceedings of Machine Learning and Systems}, 3, 2021.

\bibitem{li2020toward}
Yiran Li, Hongwei Li, Guowen Xu, Tao Xiang, Xiaoming Huang, and Rongxing Lu.
\newblock Toward secure and privacy-preserving distributed deep learning in
  fog-cloud computing.
\newblock {\em IEEE Internet of Things Journal}, 7(12):11460--11472, 2020.

\bibitem{buniatyan2019hyper}
Davit Buniatyan.
\newblock Hyper: Distributed cloud processing for large-scale deep learning
  tasks.
\newblock In {\em 2019 Computer Science and Information Technologies (CSIT)},
  pages 27--32. IEEE, 2019.

\bibitem{duan2020privacy}
Jia Duan, Jiantao Zhou, and Yuanman Li.
\newblock Privacy-preserving distributed deep learning based on secret sharing.
\newblock {\em Information Sciences}, 527:108--127, 2020.

\bibitem{zhang2019privacy}
Yitian Zhang, Hojjat Salehinejad, Joseph Barfett, Errol Colak, and Shahrokh
  Valaee.
\newblock Privacy preserving deep learning with distributed encoders.
\newblock In {\em 2019 IEEE Global Conference on Signal and Information
  Processing (GlobalSIP)}, pages 1--5. IEEE, 2019.

\bibitem{vepakomma2019reducing}
Praneeth Vepakomma, Otkrist Gupta, Abhimanyu Dubey, and Ramesh Raskar.
\newblock Reducing leakage in distributed deep learning for sensitive health
  data.
\newblock {\em arXiv preprint arXiv:1812.00564}, 2019.

\bibitem{chen2019distributed}
Jianguo Chen, Kenli Li, Qingying Deng, Keqin Li, and S~Yu Philip.
\newblock Distributed deep learning model for intelligent video surveillance
  systems with edge computing.
\newblock {\em IEEE Transactions on Industrial Informatics}, 2019.

\bibitem{balachandar2020accounting}
Niranjan Balachandar, Ken Chang, Jayashree Kalpathy-Cramer, and Daniel~L Rubin.
\newblock Accounting for data variability in multi-institutional distributed
  deep learning for medical imaging.
\newblock {\em Journal of the American Medical Informatics Association},
  27(5):700--708, 2020.

\bibitem{bae2018security}
Ho~Bae, Jaehee Jang, Dahuin Jung, Hyemi Jang, Heonseok Ha, and Sungroh Yoon.
\newblock Security and privacy issues in deep learning.
\newblock {\em arXiv preprint arXiv:1807.11655}, 2018.

\bibitem{kaissis2020secure}
Georgios~A Kaissis, Marcus~R Makowski, Daniel R{\"u}ckert, and Rickmer~F
  Braren.
\newblock Secure, privacy-preserving and federated machine learning in medical
  imaging.
\newblock {\em Nature Machine Intelligence}, 2(6):305--311, 2020.

\bibitem{zhao2020sim}
Wenshuai Zhao, Jorge {Pe{\~n}a Queralta}, and Tomi Westerlund.
\newblock Sim-to-real transfer in deep reinforcement learning for robotics: a
  survey.
\newblock In {\em 2020 IEEE Symposium Series on Computational Intelligence
  (SSCI)}, pages 737--744. IEEE, 2020.

\bibitem{zhao2020towards}
Wenshuai Zhao, Jorge {Pe{\~n}a Queralta}, Li~Qingqing, and Tomi Westerlund.
\newblock Towards closing the sim-to-real gap in collaborative multi-robot deep
  reinforcement learning.
\newblock In {\em 2020 5th International Conference on Robotics and Automation
  Engineering (ICRAE)}, pages 7--12. IEEE, 2020.

\bibitem{zhao2020ubiquitous}
Wenshuai Zhao, Jorge {Pe{\~n}a Queralta}, Li~Qingqing, and Tomi Westerlund.
\newblock Ubiquitous distributed deep reinforcement learning at the edge:
  Analyzing byzantine agents in discrete action spaces.
\newblock {\em Procedia Computer Science}, 177:324--329, 2020.

\bibitem{zhuo2019federated}
Hankz~Hankui Zhuo, Wenfeng Feng, Qian Xu, Qiang Yang, and Yufeng Lin.
\newblock Federated reinforcement learning.
\newblock {\em arXiv preprint arXiv:1901.08277}, 1, 2019.

\bibitem{nadiger2019federated}
Chetan Nadiger, Anil Kumar, and Sherine Abdelhak.
\newblock Federated reinforcement learning for fast personalization.
\newblock In {\em 2019 IEEE Second International Conference on Artificial
  Intelligence and Knowledge Engineering (AIKE)}, pages 123--127. IEEE, 2019.

\bibitem{ruan2020low}
Lihua Ruan, Sourav Mondal, Imali Dias, and Elaine Wong.
\newblock Low-latency federated reinforcement learning-based resource
  allocation in converged access networks.
\newblock In {\em Optical Fiber Communication Conference}, pages W2A--28.
  Optical Society of America, 2020.

\bibitem{nguyen2020resource}
Huy~T Nguyen, Nguyen~Cong Luong, Jun Zhao, Chau Yuen, and Dusit Niyato.
\newblock Resource allocation in mobility-aware federated learning networks: a
  deep reinforcement learning approach.
\newblock In {\em 2020 IEEE 6th World Forum on Internet of Things (WF-IoT)},
  pages 1--6. IEEE, 2020.

\bibitem{bonawitz2019towards}
Keith Bonawitz, Hubert Eichner, Wolfgang Grieskamp, Dzmitry Huba, Alex
  Ingerman, Vladimir Ivanov, Chloe Kiddon, Jakub Kone{\v{c}}n{\`y}, Stefano
  Mazzocchi, H~Brendan McMahan, et~al.
\newblock Towards federated learning at scale: System design.
\newblock {\em arXiv preprint arXiv:1902.01046}, 2019.

\bibitem{fang2020local}
Minghong Fang, Xiaoyu Cao, Jinyuan Jia, and Neil Gong.
\newblock Local model poisoning attacks to byzantine-robust federated learning.
\newblock In {\em 29th $\{$USENIX$\}$ Security Symposium ($\{$USENIX$\}$
  Security 20)}, pages 1605--1622, 2020.

\bibitem{li2020learning}
Suyi Li, Yong Cheng, Wei Wang, Yang Liu, and Tianjian Chen.
\newblock Learning to detect malicious clients for robust federated learning.
\newblock {\em arXiv preprint arXiv:2002.00211}, 2020.

\bibitem{khan2020dispersed}
Latif~U Khan, Walid Saad, Zhu Han, and Choong~Seon Hong.
\newblock Dispersed federated learning: Vision, taxonomy, and future
  directions.
\newblock {\em arXiv preprint arXiv:2008.05189}, 2020.

\bibitem{lu2020blockchain}
Yunlong Lu, Xiaohong Huang, Ke~Zhang, Sabita Maharjan, and Yan Zhang.
\newblock Blockchain empowered asynchronous federated learning for secure data
  sharing in internet of vehicles.
\newblock {\em IEEE Transactions on Vehicular Technology}, 69(4):4298--4311,
  2020.

\bibitem{wei2020federated}
Kang Wei, Jun Li, Ming Ding, Chuan Ma, Howard~H Yang, Farhad Farokhi, Shi Jin,
  Tony~QS Quek, and H~Vincent Poor.
\newblock Federated learning with differential privacy: Algorithms and
  performance analysis.
\newblock {\em IEEE Transactions on Information Forensics and Security},
  15:3454--3469, 2020.

\bibitem{zhang2020batchcrypt}
Chengliang Zhang, Suyi Li, Junzhe Xia, Wei Wang, Feng Yan, and Yang Liu.
\newblock Batchcrypt: Efficient homomorphic encryption for cross-silo federated
  learning.
\newblock In {\em 2020 $\{$USENIX$\}$ Annual Technical Conference
  ($\{$USENIX$\}$$\{$ATC$\}$ 20)}, pages 493--506, 2020.

\bibitem{deng2020adaptive}
Yuyang Deng, Mohammad~Mahdi Kamani, and Mehrdad Mahdavi.
\newblock Adaptive personalized federated learning.
\newblock {\em arXiv preprint arXiv:2003.13461}, 2020.

\bibitem{fallah2020personalized}
Alireza Fallah, Aryan Mokhtari, and Asuman Ozdaglar.
\newblock Personalized federated learning: A meta-learning approach.
\newblock {\em arXiv preprint arXiv:2002.07948}, 2020.

\bibitem{wang2019adaptive}
Shiqiang Wang, Tiffany Tuor, Theodoros Salonidis, Kin~K Leung, Christian
  Makaya, Ting He, and Kevin Chan.
\newblock Adaptive federated learning in resource constrained edge computing
  systems.
\newblock {\em IEEE Journal on Selected Areas in Communications},
  37(6):1205--1221, 2019.

\bibitem{yu2018federated}
Zhengxin Yu, Jia Hu, Geyong Min, Haochuan Lu, Zhiwei Zhao, Haozhe Wang, and
  Nektarios Georgalas.
\newblock Federated learning based proactive content caching in edge computing.
\newblock In {\em 2018 IEEE Global Communications Conference (GLOBECOM)}, pages
  1--6. IEEE, 2018.

\bibitem{kang2020training}
Jiawen Kang, Zehui Xiong, Dusit Niyato, Zhiguang Cao, and Amir Leshem.
\newblock Training task allocation in federated edge learning: A
  matching-theoretic approach.
\newblock In {\em 2020 IEEE 17th Annual Consumer Communications \&amp;
  Networking Conference (CCNC)}, pages 1--6. IEEE, 2020.

\bibitem{mohammad2019adaptive}
Umair Mohammad and Sameh Sorour.
\newblock Adaptive task allocation for asynchronous federated mobile edge
  learning.
\newblock {\em arXiv preprint arXiv:1905.01656}, 2019.

\bibitem{zhu2018broadband}
Guangxu Zhu, Yong Wang, and Kaibin Huang.
\newblock Broadband analog aggregation for low-latency federated edge learning
  (extended version).
\newblock {\em arXiv preprint arXiv:1812.11494}, 2018.

\bibitem{lu2020communication}
Yunlong Lu, Xiaohong Huang, Ke~Zhang, Sabita Maharjan, and Yan Zhang.
\newblock Communication-efficient federated learning for digital twin edge
  networks in industrial iot.
\newblock {\em IEEE Transactions on Industrial Informatics}, 2020.

\bibitem{mills2019communication}
Jed Mills, Jia Hu, and Geyong Min.
\newblock Communication-efficient federated learning for wireless edge
  intelligence in iot.
\newblock {\em IEEE Internet of Things Journal}, 7(7):5986--5994, 2019.

\bibitem{li2020talk}
Liang Li, Dian Shi, Ronghui Hou, Hui Li, Miao Pan, and Zhu Han.
\newblock To talk or to work: Flexible communication compression for energy
  efficient federated learning over heterogeneous mobile edge devices.
\newblock {\em arXiv preprint arXiv:2012.11804}, 2020.

\bibitem{mo2020energy}
Xiaopeng Mo and Jie Xu.
\newblock Energy-efficient federated edge learning with joint communication and
  computation design.
\newblock {\em arXiv preprint arXiv:2003.00199}, 2020.

\bibitem{zeng2020energy}
Qunsong Zeng, Yuqing Du, Kaibin Huang, and Kin~K Leung.
\newblock Energy-efficient radio resource allocation for federated edge
  learning.
\newblock In {\em 2020 IEEE International Conference on Communications
  Workshops (ICC Workshops)}, pages 1--6. IEEE, 2020.

\bibitem{nishio2019client}
Takayuki Nishio and Ryo Yonetani.
\newblock Client selection for federated learning with heterogeneous resources
  in mobile edge.
\newblock In {\em ICC 2019-2019 IEEE International Conference on Communications
  (ICC)}, pages 1--7. IEEE, 2019.

\bibitem{feng2020joint}
Chenyuan Feng, Yidong Wang, Zhongyuan Zhao, Tony~QS Quek, and Mugen Peng.
\newblock Joint optimization of data sampling and user selection for federated
  learning in the mobile edge computing systems.
\newblock In {\em 2020 IEEE International Conference on Communications
  Workshops (ICC Workshops)}, pages 1--6. IEEE, 2020.

\bibitem{lu2020privacy}
Xiaofeng Lu, Yuying Liao, Pietro Lio, and Pan Hui.
\newblock Privacy-preserving asynchronous federated learning mechanism for edge
  network computing.
\newblock {\em IEEE Access}, 8:48970--48981, 2020.

\bibitem{lu2019differentially}
Yunlong Lu, Xiaohong Huang, Yueyue Dai, Sabita Maharjan, and Yan Zhang.
\newblock Differentially private asynchronous federated learning for mobile
  edge computing in urban informatics.
\newblock {\em IEEE Transactions on Industrial Informatics}, 16(3):2134--2143,
  2019.

\bibitem{zhang2020poisongan}
Jiale Zhang, Bing Chen, Xiang Cheng, Huynh Thi~Thanh Binh, and Shui Yu.
\newblock Poisongan: Generative poisoning attacks against federated learning in
  edge computing systems.
\newblock {\em IEEE Internet of Things Journal}, 2020.

\bibitem{ferrer2018blockchain}
Eduardo~Castell{\'o} Ferrer.
\newblock The blockchain: a new framework for robotic swarm systems.
\newblock In {\em Proceedings of the future technologies conference}, pages
  1037--1058. Springer, 2018.

\bibitem{queralta2019blockchain}
Jorge {Pe\~na Queralta} and Tomi Westerlund.
\newblock Blockchain-powered collaboration in heterogeneous swarms of robots.
\newblock {\em arXiv preprint arXiv:1912.01711}, 2019.

\bibitem{nawaz2019edge}
Anum Nawaz, Tuan~Nguyen Gia, Jorge {Pe\~na Queralta}, and Tomi Westerlund.
\newblock Edge ai and blockchain for privacy-critical and data-sensitive
  applications.
\newblock In {\em 2019 Twelfth International Conference on Mobile Computing and
  Ubiquitous Network (ICMU)}, pages 1--2. IEEE, 2019.

\bibitem{nawaz2020edge}
Anum Nawaz, Jorge {Pe{\~n}a Queralta}, Jixin Guan, Muhammad Awais, Tuan~Nguyen
  Gia, Ali~Kashif Bashir, Haibin Kan, and Tomi Westerlund.
\newblock Edge computing to secure iot data ownership and trade with the
  ethereum blockchain.
\newblock {\em Sensors}, 20(14):3965, 2020.

\bibitem{queralta2020blockchain}
Jorge {Pe\~na Queralta} and Tomi Westerlund.
\newblock Blockchain for mobile edge computing: Consensus mechanisms and
  scalability.
\newblock {\em arXiv preprint arXiv:2006.07578}, 2020.

\bibitem{nguyen2021federated}
Dinh~C Nguyen, Ming Ding, Quoc-Viet Pham, Pubudu~N Pathirana, Long~Bao Le,
  Aruna Seneviratne, Jun Li, Dusit Niyato, and H~Vincent Poor.
\newblock Federated learning meets blockchain in edge computing: Opportunities
  and challenges.
\newblock {\em arXiv preprint arXiv:2104.01776}, 2021.

\bibitem{bao2019flchain}
Xianglin Bao, Cheng Su, Yan Xiong, Wenchao Huang, and Yifei Hu.
\newblock Flchain: A blockchain for auditable federated learning with trust and
  incentive.
\newblock In {\em 2019 5th International Conference on Big Data Computing and
  Communications (BIGCOM)}, pages 151--159. IEEE, 2019.

\bibitem{majeed2019flchain}
Umer Majeed and Choong~Seon Hong.
\newblock Flchain: Federated learning via mec-enabled blockchain network.
\newblock In {\em 2019 20th Asia-Pacific Network Operations and Management
  Symposium (APNOMS)}, pages 1--4. IEEE, 2019.

\bibitem{otoum2020blockchain}
Safa Otoum, Ismaeel Al~Ridhawi, and Hussein~T Mouftah.
\newblock Blockchain-supported federated learning for trustworthy vehicular
  networks.
\newblock In {\em GLOBECOM 2020-2020 IEEE Global Communications Conference},
  pages 1--6. IEEE, 2020.

\bibitem{chai2020hierarchical}
Haoye Chai, Supeng Leng, Yijin Chen, and Ke~Zhang.
\newblock A hierarchical blockchain-enabled federated learning algorithm for
  knowledge sharing in internet of vehicles.
\newblock {\em IEEE Transactions on Intelligent Transportation Systems}, 2020.

\bibitem{pokhrelfederated}
Shiva~Raj Pokhrel and Jinho Choi.
\newblock Federated learning with blockchain for autonomous vehicles: Analysis
  and design.

\bibitem{gaofeng2020blockchain}
Hua Gaofeng, Li~Zhu, Jinsong Wu, Chunzi Shen, Lingyan Zhou, and Qingqing Lin.
\newblock Blockchain-based federated learning for intelligent control in heavy
  haul railway.
\newblock 2020.

\bibitem{pokhrel2020federated}
Shiva~Raj Pokhrel.
\newblock Federated learning meets blockchain at 6g edge: A drone-assisted
  networking for disaster response.
\newblock In {\em Proceedings of the 2nd ACM MobiCom Workshop on Drone Assisted
  Wireless Communications for 5G and Beyond}, pages 49--54, 2020.

\bibitem{goldberg2013cloud}
Ken Goldberg and Ben Kehoe.
\newblock Cloud robotics and automation: A survey of related work.
\newblock {\em EECS Department, University of California, Berkeley, Tech. Rep.
  UCB/EECS-2013-5}, 2013.

\bibitem{savazzi2021opportunities}
Stefano Savazzi, Monica Nicoli, Mehdi Bennis, Sanaz Kianoush, and Luca
  Barbieri.
\newblock Opportunities of federated learning in connected, cooperative and
  automated industrial systems.
\newblock {\em arXiv preprint arXiv:2101.03367}, 2021.

\bibitem{tanwani2019fog}
Ajay~Kumar Tanwani, Nitesh Mor, John Kubiatowicz, Joseph~E Gonzalez, and Ken
  Goldberg.
\newblock A fog robotics approach to deep robot learning: Application to object
  recognition and grasp planning in surface decluttering.
\newblock In {\em 2019 International Conference on Robotics and Automation
  (ICRA)}, pages 4559--4566. IEEE, 2019.

\bibitem{li2019fc}
Zhaoran Li, Lujia Wang, Lingxin Jiang, and Cheng-Zhong Xu.
\newblock Fc-slam: Federated learning enhanced distributed visual-lidar slam in
  cloud robotic system.
\newblock In {\em 2019 IEEE International Conference on Robotics and
  Biomimetics (ROBIO)}, pages 1995--2000. IEEE, 2019.

\bibitem{zhang2021distributed}
Zijian Zhang, Shuai Wang, Yuncong Hong, Liangkai Zhou, and Qi~Hao.
\newblock Distributed dynamic map fusion via federated learning for intelligent
  networked vehicles.
\newblock {\em arXiv preprint arXiv:2103.03786}, 2021.

\bibitem{majcherczyk2020flow}
Nathalie Majcherczyk, Nishan Srishankar, and Carlo Pinciroli.
\newblock Flow-fl: Data-driven federated learning for spatio-temporal
  predictions in multi-robot systems.
\newblock {\em arXiv preprint arXiv:2010.08595}, 2020.

\bibitem{liu2020federated02}
Boyi Liu, Lujia Wang, Ming Liu, and Cheng-Zhong Xu.
\newblock Federated imitation learning: A novel framework for cloud robotic
  systems with heterogeneous sensor data.
\newblock {\em IEEE Robotics and Automation Letters}, 5(2):3509--3516, 2020.

\bibitem{busart2020federated}
Carl~E Busart~III.
\newblock {\em Federated Learning Architecture to Enable Continuous Learning at
  the Tactical Edge for Situational Awareness}.
\newblock PhD thesis, The George Washington University, 2020.

\bibitem{liang2019federated}
Xinle Liang, Yang Liu, Tianjian Chen, Ming Liu, and Qiang Yang.
\newblock Federated transfer reinforcement learning for autonomous driving.
\newblock {\em arXiv preprint arXiv:1910.06001}, 2019.

\bibitem{liu2019lifelong}
Boyi Liu, Lujia Wang, and Ming Liu.
\newblock Lifelong federated reinforcement learning: a learning architecture
  for navigation in cloud robotic systems.
\newblock {\em IEEE Robotics and Automation Letters}, 4(4):4555--4562, 2019.

\bibitem{papadopoulos2020towards}
Georgios~Th Papadopoulos, Margherita Antona, and Constantine Stephanidis.
\newblock Towards open and expandable cognitive ai architectures for
  large-scale multi-agent human-robot collaborative learning.
\newblock {\em arXiv preprint arXiv:2012.08174}, 2020.

\bibitem{papadopoulos2021user}
Georgios~Th Papadopoulos, Asterios Leonidis, Margherita Antona, and Constantine
  Stephanidis.
\newblock User profile-driven large-scale multi-agent learning from
  demonstration in federated human-robot collaborative environments.
\newblock {\em arXiv preprint arXiv:2103.16434}, 2021.

\end{thebibliography}

\end{document}